\documentclass[letterpaper]{article}
\usepackage{aaai}
\usepackage{times}
\usepackage{graphicx}
\usepackage{helvet}
\usepackage{courier}
\usepackage[utf8]{inputenc}
\begin{document}
%
\title{Using reinforcement learning to design an AI assistant\\ for a satisfying co-op experience {}}
\author{Ajay Krishnan, Niranj Jyothish, Xun Jia\\
NC State University \\
}
\maketitle
\begin{abstract}
\begin{quote}
In this project, we designed an intelligent assistant player for a single-player game Space Invaders with the aim to provide a satisfying co-op experience.  The agent behavior was designed using reinforcement learning techniques and evaluated based on several criterion. We validate the hypothesis that an AI driven computer player can provide a satisfying co-op experience.  
\end{quote}
\end{abstract}

\noindent 

\section{Introduction}
Artificial Intelligence has always been an integral part of video games ever since their inception. Ever since the 1950s AI has been a part of video games interacting with the players as a competitor, a supporter or working in the background invisible to the player molding their gaming experience. Thus video game AI differs from Academic AI in that the primary aim is to improve the players gaming experience. In the 1950s AI took the form of computer programs written to play games such as Chess and Checkers. Work on these programs eventually led to the IBM Deep Blue defeating Garry Kasparov a world champion. Games such as Pacman (198), Castle Wolfenstein (1981) are examples of where the AI controlled all the enemies on the screen and was programmed to challenge the player. Sid Mier's Civilization (1991) had AI with which the player would be able to engage in diplomacy forming alliances and going to war.  AI has also been used in interactive simulations wherein users chose to take actions and AI dynamically creates scenarios based on that. It has been used in games such as SIMS(2000), Lemmings(1991). AI has taken the form of a story-teller too in games like Face (2005) and Skyrim (2011). Games like Minecraft(2009) and No Man's Sky(2016) are great examples of Procedural Content Generation. Followers, companions and allies in some of the games mentioned above also show the co-operative uses of AI in games. 

  A number of data structures and algorithms are in use to implement these use-cases of AI. Finite state machines(FSM) algorithm is a relatively simple A.I. where designers create a list of all possible events a bot can experience. The designers then assign specific responses the bot would have to each situation. (Wolfenstein). A more standard approach of Monte Carlo Search Tree can  also be used to prevent the repeatability of FSM. It visualizes all possible moves from a bot’s current position and then for each move, it considers all of its responding moves it could make. Another popular data structure is behavior trees.  It is a model of plan execution that is graphically represented as a tree. A node in a tree either encapsulates an action to be performed or acts as a control flow component that directs traversal over the tree. Thus what algorithm the AI is based on plays an important role in the final game experience. This project focuses on applying one such algorithm to produce co-operative AI providing better experience.
  
\subsection{Background}

The aim of the research project is to develop an AI assistant that accompanies the player character. The presence of a second player in games provide a lot of dynamics to the users game experience. A second player allows more freedom in level design for game developers allowing the design of harder levels which would be a struggle or impossible if attempted to be solved with a single player. Many co-op games have a second player connecting to a primary players session with the aim of assisting him in return for some reward. The most popular examples would be from the dark souls and monster hunter series, where players join other player's game sessions to help assist fighting a boss and once finished get a share of the loot. In absence of a multiplayer framework or a second player to play, a simple rule based bot would server to recreate the above mentioned experience. However rule-based systems are rigid and an AI that is trained to play the game like a human would provide a better co-op experience. Better AI based agents also opens opportunities where the AI agent can behave as a contingency for multiplayer games in case of network failures where the agent takes control of the disconnected player character thus allowing perhaps both players to continue their game session without sacrificing a satisfying game experience. Both Dark Souls and Monster Hunter World makes the enemies tougher when fighting them as a group. However if your ally disconnects the enemy is not scaled back down and instead the player is left to deal with a much stronger enemy alone affecting their enjoyment. We will be exploring the use of reinforcement learning techniques like Deep Q-learning to implement the AI assistant. In reinforcement learning the agent learns by interacting with the environment through trial and error. The agent receives rewards for actions it performs based on the state of the environment and the goal is to maximize the expected cumulative reward over time. The technique has had major breakthroughs in games such as Go, Starcraft 2 and Dota 2 where they are able to perform to a very considerable high skill level. The specific technique we are looking into Deep Q-learning is a variant of Q-learning that uses a neural network to overcome the problems Q-learning faces with scalability. We applied our approach to a python implementation of the classic space invader game modified as a two player version. Deep Q-learning has already been shown to work on the original Atari 2600 Space Invader. However training an agent that serves as an assistant to the first player makes the environment a multi-agent where the first player is neutral to the second player but the second player has to co-operate with the first player. Our goal is to check whether Deep Q-learning can be implemented in such an environment.
\section{Related Work Review}

Google's deep mind has made several breakthroughs in the field of AI through many different reinforcement learning. The Deep Q-Learning algorithm manages to develop AI that can play many Atari games such as Space Invaders, Pong, Breakout, etc. to an almost expert level [1]. While the above mentioned games are all 2D games. Deep Reinforcement Learning has also been tested in 3D environments, specifically the Asynchronous Actor-Critic algorithm, combines Deep Q-Networks with a deep policy network for selecting actions [2]. The possibilities that Deep Reinforcement Learning provides where only further highlighted with the development of an algorithm for the classic board game Go that combined tree search for evaluation positions and a deep neural network for selecting moves [3]. The playing with Atari Deep Reinforcement Learning paper serves as the the inspiration and the base for this project.

Reinforcement learning in multi agent settings has been already tested in multi agent environments in which all the agents work towards a common goal [5]. In the above method the agents learn separately, but after a set number of iterations distribute the policy learned to its allies. The system we are considering however the first player behaves independently of the assistant. The assistant tries to learn in this environment by considering the player to be a part of the environment.

We have also found similar projects that aim at building assistant bots within games. CraftAssist[7] explores the implementation of a bot assistant in Minecraft. The bot appears like a human player, does its own thing moving around and modifying the world and can be communicated to using in-game chat.
\section{Algorithm Overview - Deep Q-Learning}
  
  Reinforcement learning is an area of machine learning concerned with how software agents ought to take actions in an environment in order to maximize some notion of cumulative reward. The problems of interest in reinforcement learning have also been studied in the theory of optimal control, which is concerned mostly with the existence and characterization of optimal solutions, and algorithms for their exact computation, and less with learning or approximation, particularly in the absence of a mathematical model of the environment. In economics and game theory, reinforcement learning may be used to explain how equilibrium may arise under bounded rationality.

  Q-learning is a model-free reinforcement learning algorithm. The goal of Q-learning is to learn a policy in the sense that it maximizes the expected value of the total reward over any and all successive steps, starting from the current state The policy tells an agent what action to take under what circumstances. It does not require a model of the environment, and it can handle problems with stochastic transitions and rewards, without requiring adaptations. In Q-Learning, the agent interacts with the environment iteratively, by taking an action. The environment responds by informing the agent of the reward from that action, and advancing to the next state. This happens continuously until the environment is “solved”. From a positive reward the agent learns that the action was suitable for the state it was in and a negative reward teaches that it was not. Reinforcement learning tries to learn the best sequence of actions to take. This is done by trying different combinations of actions, first randomly, than using a policy based on what the model has learned from rewards up to this point. This happens until the environment reaches its terminal state.
  
  For each iteration of Q-learning we consider the following:\\
   a - action taken by the agent onto the environment at timestep t\\
   s - state held in the environment at timestep t\\
   r - state held in the environment at timestep t\\
   $\pi$ - policy used by the agent to decide the next action\\
   Q - the long term return for the agent, when it takes action a at state s. Unlike r, which is the short term reward, Q refers to the combined reward of all future states and actions, starting from its current position.\\
  The goal of Q-learning is to learn these Q-mappings for each state-action pair. Over multiple iterations, the policy used to decide the next action is improved, taking in account which action returns the highest future reward at that state. Figure 1 shows a the Q-learning pseudo code and Figure 2 explains the Bellman equation used to update the Q-function in more detail.
  
  Learning rate \(\alpha\) : The learning rate or step size determines to what extent newly acquired information overrides old information. A value of 0 makes the agent learn nothing (exclusively exploiting prior knowledge), while a value of 1 makes the agent consider only the most recent information (ignoring prior knowledge to explore possibilities). In fully deterministic environments \(\alpha\) is 1, while in stochastic environments, the algorithm converges under some technical conditions on the learning rate that require it to decrease to zero. In practice a constant learning rate is used for all time steps.
  
  Discount rate \(\gamma\) : The discount rate determines the importance of future rewards. A value of 0 makes the agent short-sighted and consider only the immediate reward which is R(s,a) in the equation shown above. A value that is closing towards 1 will make it work towards a long-term high reward. A value greater than or equal to 1 may make the action values diverge.  
  
  \begin{figure}
    \includegraphics[width=8cm, height=2cm]{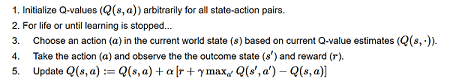}
    \caption{Q-Learning pseudo-code}
\end{figure}   

\begin{figure}
    \includegraphics[width=8cm, height=2cm]{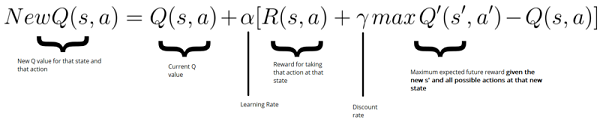} 
    \caption{Bellman equation}
\end{figure}

  Q-learning itself is only suitable for small environments and suffers when the number of states and actions in the environment increases since producing and updating the Q-table (The table of Q-mappings) become ineffective. Because of this we will be using a variant of Q-learning developed by Google Deep Mind called Deep Q-learning. Deep Mind proved the success of Deep Q-learning techniques by showing the algorithm playing Atari 2600 games at an expert human level. As mentioned before our test bed space invaders is one of those Atari 2600 games that Deep Q-learning works for. 

\begin{figure}
    \includegraphics[width=9cm, height=6cm]{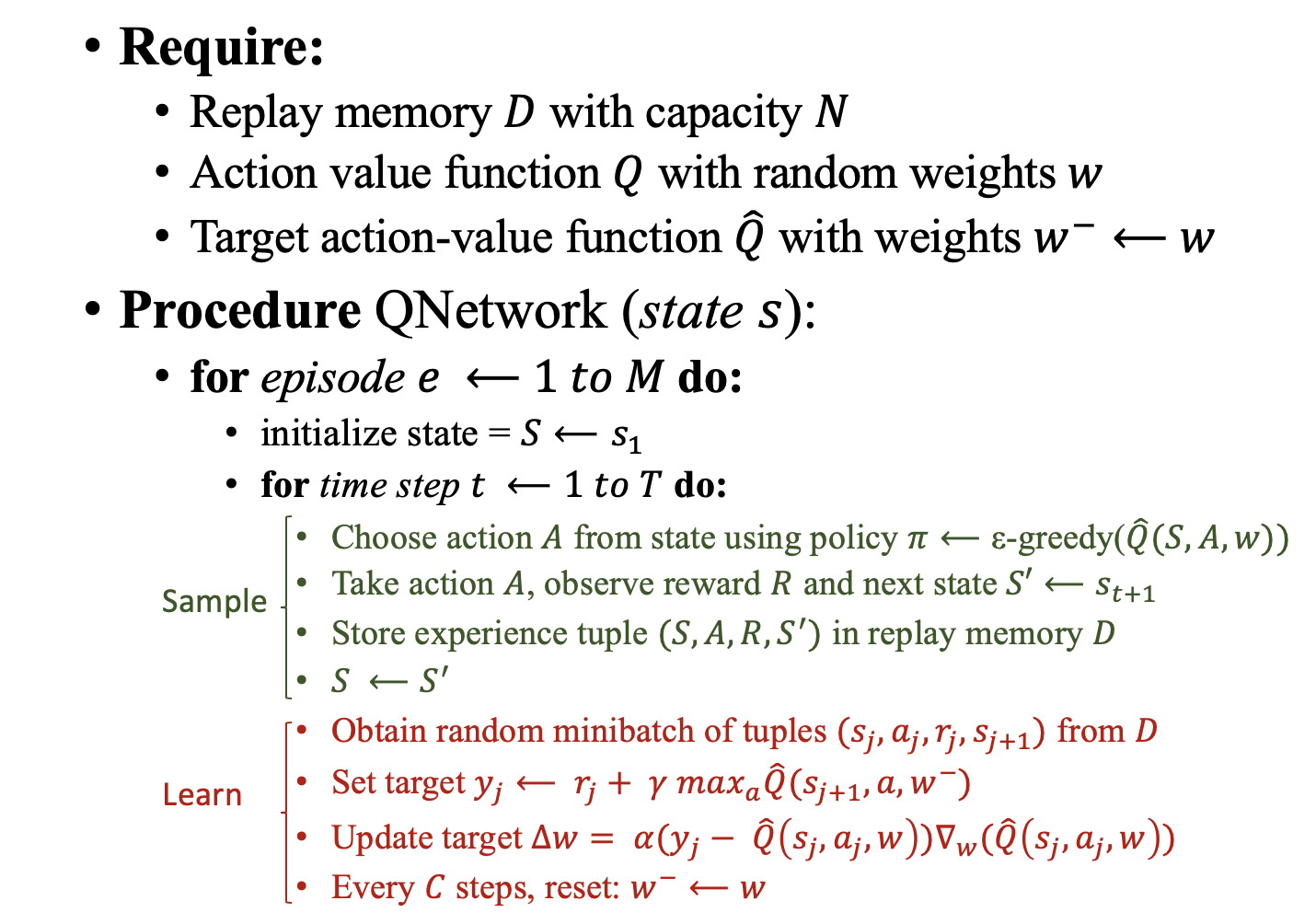}
    \caption{Deep Q-learning pseudocode}
\end{figure}    

\begin{figure}
    \includegraphics[width=9cm, height=4cm]{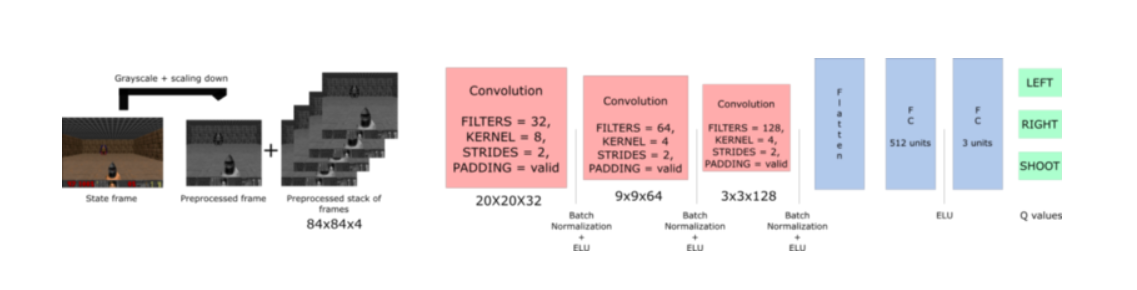}
    \caption{Deep Q-learning architecture}
\end{figure}

The use of convolution neural networks a subclass of deep neural networks have has made major breakthroughs in computer vision and automatic speech recognition. Using sufficiently large data sets, convolution neural networks even with little prepossessing compared to other image classification algorithms learns representations better than in traditional handcrafted algorithms. It is this success that motivated the development of Deep Q-learning. By combining the Q-learning reinforcement learning algorithm with deep neural networks  that operate directly on RGB images and efficiently processing training data using stochastic gradient descent we obtain Deep Q-learning. In Deep Q-learning instead of tracking the Q-mappings for each state action pair in a Q-table we use the convectional neural network to learn the Q-value function instead. Figure 3 shows the pseudo code for Deep Q-learning implemented by Deep Mind. There are two training steps in each iteration. The first step uses the epsilon-greedy policy to decide and take an action based on the current state. The second step involves a "Replay Buffer". Each seen state/action/reward/next-action pair is stored in a dictionary called the Replay buffer. A small batch of these is sampled, and we use them to update the local Q Network. There are a few advantages for using Experience Replay buffer. First, each step of the experience is potentially used in many weight updates thus providing greater data efficiency. Secondly, it prevents the Neural Network from relying on sequential experiences for learning. Experiences in an environment can be highly correlated, and training the model with a random sample breaks the temporal correlation by using independently distributed data. This prevents the model from oscillating between states, and it easier for the model to converge.

The Q-value function that we use is obtained as mentioned before using a neural network. We use an architecture in which there is a separate output unit for each possible action, and only the state representation is an input to the neural network. The outputs correspond to the predicted Q-values of the individual action for the input state. Figure 4 shows an example architecture for deep Q-learning implemented for the game Doom. Initial step is to pre-process the input state to reduce complexity. The raw frame is converted to gray scale since RGB information is not necessary for that particular implementation. The frame is then cropped to consider only important play area. A stack of 4 sub frames is used as input to solve the problem of temporal limitation where the agent wouldn't be able to perceive the target direction of motion from a single frame. The frames are then processed by three convolution layers. The final output layer provides an output Q-value for each possible action. The convolution networks trained using this approach is referred to as Deep Q-Networks. 

\section{Environment Overview}

\subsection{Space Invaders}
\begin{figure}
    \includegraphics[width=8cm, height=6cm]{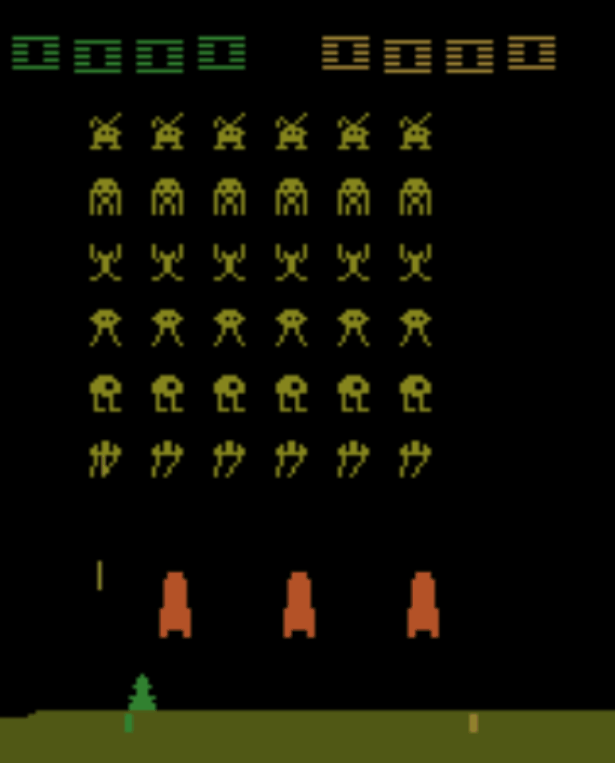}
    \caption{Original Atari 2600 Space Invaders game for comparison of difficulty}
\end{figure} 

  Space Invaders is a game which was released in 1978 and soon became the most influential game at that time. The aim is to defeat five rows of aliens that move horizontally back and forth across the screen as they advance toward the bottom of the screen. The player is partially protected by several stationary defense bunkers that are gradually destroyed from the top and bottom by blasts from either the aliens or the player. The player defeats an alien and earns points by shooting it with the laser cannon. As more aliens are defeated, the aliens' movement and the game's music both speed up. Defeating all the aliens on-screen brings another wave that is more difficult, a loop which can continue endlessly. A special "mystery ship" will occasionally move across the top of the screen and award bonus points if destroyed. The aliens attempt to destroy the player's cannon by firing at it while they approach the bottom of the screen. If they reach the bottom, the alien invasion is declared successful and the game ends tragically; otherwise, it ends generally if the player is destroyed by the enemy's projectiles and the player has no lives left.

\begin{figure}
    \includegraphics[width=8cm, height=6cm]{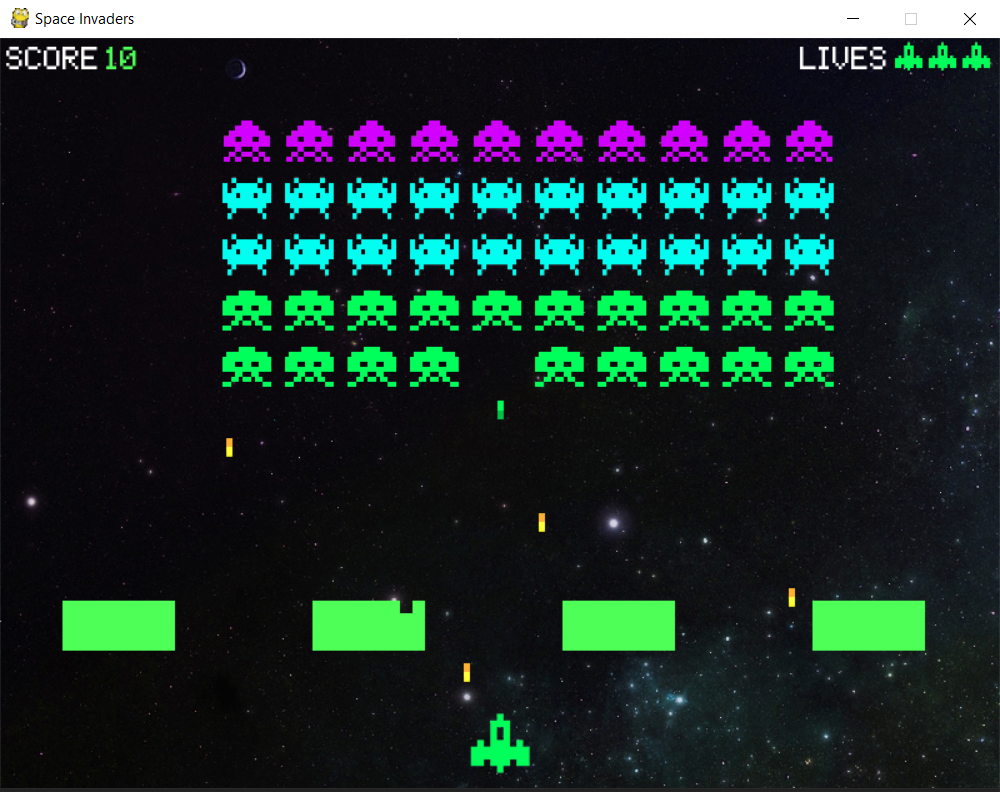}
    \caption{Single player space invader game}
\end{figure} 

The original Atari 2600 Space Invader game is a single player game and very easily winnable by a single player and a second player would serves no purpose. The python implementation used for this research project however has more enemies, move faster and shoot faster making the game more difficult to win and provides a reason for an assistant to be added to make the game easier. The enemies in this implementation of the game do not track the player characters and shoot, but instead shoot randomly. The first player has 5 lives, losing all of them leads to a game over. Figure 5 shows the original game compared to the implementation we are working on Figure 6 which has more enemies justifying the addition of a second player. The second player added to the game is unaffected by enemy fire, however still loses the game if the first player dies. This allows for easier training of the second agent since it need not be concerned with survival. The simplicity of the game is seen in its small action set with the player only being able to move left, move right or shoot at each frame. The small size of the action set means the number of actions that the agent has to choose from at each observed state is also small. The simple pixel of the game without any complicated graphics also  makes the game highly suitable as a test environment for deep Q-learning which only observes the current image of the game screen to learn. 
 
 \begin{figure}
    \includegraphics[width=8cm, height=6cm]{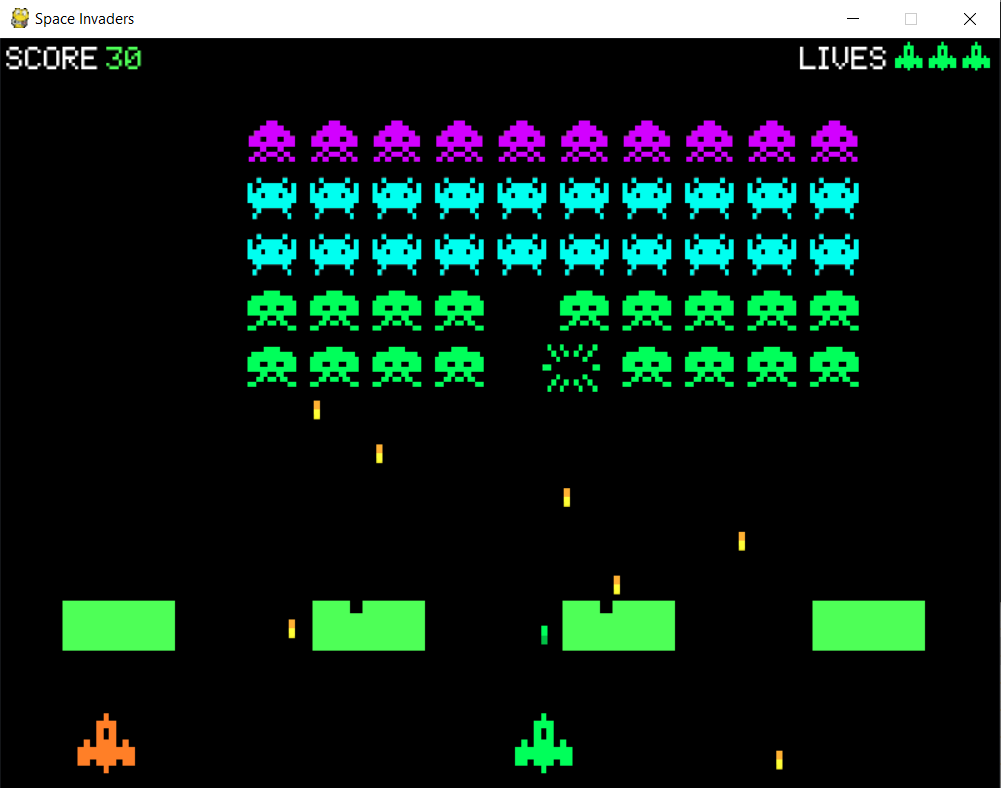}
    \caption{Two-player space invader game}
\end{figure}  

\subsection{PyGame Learning Environment}
  The original paper on which this project is based on used Atari 2600 ROMS that were run using an emulator. The ROM is emulated using the Open AI gym which wraps the game and makes it a suitable environment for applying learning algorithms. However since the implementation of the game we are using is written in Python using Pygame we have instead chosen to use the PyGame Learning Environment(PLE) as a wrapper for the game which is much more compatible and easier to implement. The functionalities that PLE provides us can be described by the following functions:
  
  init() - Resets the game to its initial state which we do so every time the game enters a done state.
  
  act() - Passes an action to the game and the game progresses 1 frame based on that action and returns the reward associated with that action.
  
  getScreenRGB() - Provides an array of the RGB values of the pixels on screen at the current state.
  
  There are three actions possible for the agents to perform. Move Left, Move right and Shoot. PLE allows only one action to be send to the wrapped game. However for the two player version of the game that we are using as test environment we need to pass actions for both players. Hence we modify the game to accept a value that represents the combination of actions performed int the form of LL, LR, LS, RL, RR, RS, SL, SR, SS, where L refers to Move Left, R refers to Move Right and S refers to Shoot.

\section{Experiments}

\subsection{Preparing Experiments}

We began by making two versions of our test environment the Space Invader game, a single-player version and a two-player version which are shown in Figure 6 and Figure 7 respectively. We reconstructed a Deep Q-Network for our game implementation based on the Deep Mind's idea of Deep Q-learning to develop an agent that simply plays the game to the best of its ability. 

    We provide as input for the algorithm an array of RGB values of the pixels on screen for each frame to the neural network. Feeding only one image-frame gives us no more than the position of the player ship, the enemies and the missiles. We are left with no information regarding velocity or acceleration. A choice of two frames only gives us the velocity. Experimentally, it has been found that, feeding four consecutive frames of the game state gives the most amount of information to the agent while still being cost-effective in terms of performance. Thus we provide as input to the Deep Q-network a stack of four frames as input. The network then provides as output a vector of Q-values for each action possible by the agent. While in exploitative mode we take the action corresponding to the biggest Q-value we obtained. 
    
    The agent switches between exploitation and exploration (choosing an action randomly) based on a parameter epsilon. We give a high value to epsilon at the start of the training to incentivize exploration, but over time it decreases so that the learned information can be exploited. Once the action has been decided on we passed it to the wrapped game to advance a frame and obtain the next state. We also obtain through the PLE framework the reward for that action. We assigned a positive reward of +30 any time an enemy ship is destroyed, a negative reward of -10 every time the player loses a life and -10 when the game is over. Based on these rewards, the learning rate and the discount rate the Q-network weights are corrected. Thus over time the agent identifies best actions associated with each state and the network converges to an optimal sequence of actions.   

As mentioned in the algorithm overview one of the main problems with Deep Q-learning is the stability. When we are feeding observations into our network, the network generalizes over the recent past experiences. Thus, comes in the concept of experience replay. To avoid this problem, we record the agent’s action-observation into an experience tuple (State, Action, Reward, Next State) in replay memory and randomly sample from these experiences. This helps the agent to learn what actions to take given an observation and helps to stabilize the training.  The Q-learning updates are incremental and do not converge quickly, so multiple passes with the same data is beneficial, especially when there is low variance in immediate outcomes (reward, next state) given the same state, action pair. We save the weights of the network at major checkpoints determined by the number of sessions of the game for which training has occurred. This allowed us to pause and continue the training as well as use the Q-network for the next step. 
 
The main purpose of training the single player agent was to use it to train the second player. Due to the time as well as the large number of sessions over which we train, it is impossible to train the assistant using a human to control the first player. One option would have been to make the first player perform random actions. However doing so would mean that the player is likely to die very quickly due to the onslaught of missiles fired by the enemy or get stuck at the corner of the screen without attacking any enemies. However using a trained agent as a substitute for the human player overcomes all this. The agent allows us to train for as long as we want and the game sessions are also longer than they were when a random agent was used. Longer game sessions mean that during training the assistant agent is able to reach new  and existing states more frequently thus learning more about the actions to perform and converging faster.

\subsection{Training}
Similar to training the single player agent, we began by wrapping the two player implementation with the PLE framework and setting up the environment as mentioned in the environment overview section. This time however we maintain two Deep Q-Networks. We assign weights to the first Q-Network by restoring from the final checkpoint we made while training the single player agent. We will use this network to derive the actions of the first player. The second Q-Network will be setup similar to how the Q-Network was setup during the training of the single player agent.
The training of the second Q-Network will follow mostly the exact steps as during the single-player training. The only difference is that this time we pass the processed stack of frames to both Q-Networks. There is no exploration for the first Q-Network and we obtain the action which the network thinks provides the best outcome. Once we've obtained both actions they are combined and send to the game to get the next state and obtain the reward to train the assistant.

The rewards that we assign for the assistant agent are done so with the aim of making the agent useful to the player. The assistant itself does not have lives but instead will be assigned a negative reward when the player dies. Similarly a game over also gives a massive negative reward. The positive reward for shooting down the enemy ships is calculated differently. We calculate the horizontal distance between the player ship and the enemy that was shot down. We map this value to a range 0 - 1 with 1 being the ship the farthest from the player. The positive reward to be returned is multiplied with this value. This is to incentivize the agent to focus on the enemies farther from the player. However once the player goes below 3 lives the above value is subtracted from 1 and then multiplied with the positive reward thus focusing on enemies closer to the player. Similar to the single player agent we save the weights of the assistant network at regular checkpoints.

\begin{center}
\begin{tabular}{ |c|c| } 
 \hline
 80 & Game win  \\ 
 kill score * 50 & Assistant kill based on player lives left \\ 
 -20 & Life loss  \\ 
 \hline
\end{tabular}
\end{center}
if player lives $>$ 3:
\begin{equation}
kill score = |player.position - enemy.position| 
\end{equation}
if player lives $<$ 3:
\begin{equation}
    kill score = 1- |player.position - enemy.position| 
\end{equation}
Once finished training, we modified the environment so that instead of using the Q-Network to drive the player, we obtain keyboard inputs which we append with the assistant action and pass to the game. This allows us to play the game and test how the assistant behaves according to our actions which differ from the trained player agent. Initially we planned to rate the agent on the through play testing  and a follow up survey. Our evaluation criteria was based on 3 aspects. The first is “Is it helpful?”, the second is “Is it human-like?” and the third is “Is it like a player or is it like an assistant?”. That is to say, a successful assistant will cover more tasks if the player is behaving weak, and vice versa. At the same time, it should behave like an assistant, a cooperator instead of a competitor which competes with the player to get higher scores. What's more, it has to be human-like, which means it should move and shoot with a "purpose" instead of hanging around and shooting stochastically. However limiting circumstances have made the above tests difficult to setup and we had to choose a different evaluation method. We now have collected two sets of data. The first is the data of the final scores for the single-player agent (Figure 8) . This serves to show the improvement a second player brings to the game. Second we have a two player agent where the second player does random actions (Figure 10). And finally we have the training data for the assistant AI over more than 1500 games (Figure 12). 
 
 \begin{figure}
    \includegraphics[width=8cm, height=6cm]{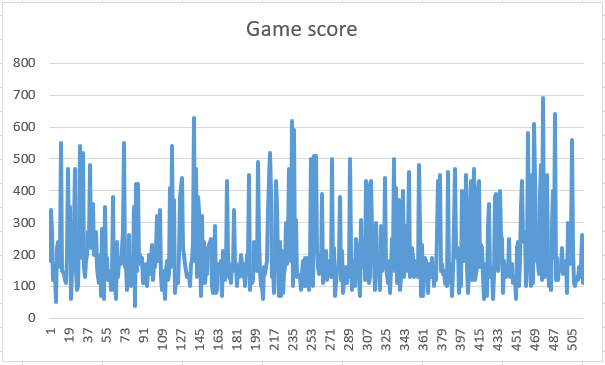}
    \caption{Single Player Scores}
\end{figure}

\section{Results and Observations}

Some preliminary information regarding the game-play is necessary. We have noted that the game winning scenarios end up having a score close to 2000. We are not able to provide a fixed threshold above which we can consider the game to be won, because of the presence of the mystery ship which is worth more. However scores close to 2000 being assumed as a winning scenario is a reasonable assumption for analysing our available data. Moreover it is hard to evaluate the efficiency of the reinforcement learning agent based on the final scores because as it is shown in the graph the results vary a lot. This is because small changes in weight can cause huge changes in how the agent behaves. 

\begin{figure}
    \includegraphics[width=8cm, height=6cm]{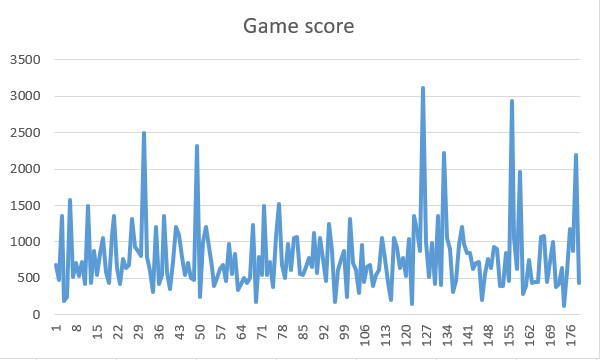}
    \caption{Two player implementation with random assistant scores}
\end{figure}

The single player training data shows an average score close to 200 , the max reward observed is approximately 700. This data will provide a good baseline to compare our two player implementations against, mainly showing the necessity of the two-player implementation. 

The two-player implementation using random actions has an average score close to 800 , the max reward observed is above 3000. The training data for the two-player implementation with assistant AI is gathered over more than 1500 games, we see that the average score converges to a value bit more than 600, the maximum score observed during the training is 2500. An initial look would suggest that the trained agent under performs in comparison to just performing random actions. However the graph data only shows us the overall effectiveness of the agent from a score perspective. We require observational data to conclude the agents performance for an ideal experience. We were also able to find a couple of players to play the game and collected their opinions based on a few criteria. We make the testers play the single-player version of the game first so that they have a base line with which they can compare the experience of using an assistant. The players gave their opinions on four criteria:

i)   How helpful the agent was?

ii)  Whether the agent felt random or acted with purpose?

iii) Whether the agent felt more like an assistant or as a competitor?

iv)  How the assistant affected the overall game experience?

\section{Analysis}

From the visualized data of the two-player implementations, we interpreted the following:

I) The random agent while occasionally receiving a higher score than the trained agent plays very unnaturally and would obviously serve no purpose as an assistant in a co-op game. However, the trained agent does behave with some level intelligence being able to properly identify and shoot down aliens in states it has thoroughly explored.

II) It was observed that while the single-player implementation of the agent actively played the game dodging bullets and taking down enemies on a consistent basis. The two-player implementation would in some games simply move to the left of the screen and stay there taking shots at the aliens as they move in front of it. We conclude this is due to a lack of training in those specific states. With two player objects being considered by the network for the assistant agent, there are a larger number of states for it to consider, which means much more training than in the case of the single-player agent. However it is to be noted that when the agent does this it still gets rewarded due to the base game we have built on. In our space invaders implementations all the aliens move from left to right. Even if the column of aliens at the edges are wiped out, the next column will start hitting the edge of the screen. This means that even an agent that sits at the edge of the screen and fires will still maximize score. Since time is not a concern in the space invader game, there is no incentive to chase the aliens and shoot them. This leads to our next observation.

III) The optimal actions that the agent ends up learning is affected by what the environment supports and rewards. This is what led to the above mentioned scenario. In a dynamic environment like Space Invaders it becomes very obvious how inhuman-like it is for an agent to behave that way. We see more activity from the agents during the training than when it only follows the optimal policy, this is mainly due to the exploration part of training. Thus it might be preferable to use an epsilon-greedy policy when implementing the trained agent, just so that the agent doesn't get stuck in states where it isn't experienced enough. 

IV) While the results do show that reinforcement learning provides results it is important to consider how we actually apply this usefully within the game industry. We argue that reinforcement learning is best applied where it is the strongest. It is not practical or a good investment to spend time training agents for action such as following the player or for performing specific action sequences dynamically. Instead it is best used in repeating game-play scenarios, where the bounds of the environment can be clearly defined. Such as a combat scenario which is different from the open-world game-play, an example would be Final Fantasy or Breath of Fire where the fights take place with a different interface and environment compared to the rest of the game. It need not be turn-based but needs to be sufficiently controllable for the agent to learn efficiently. Very large unbounded environments would mean an unimaginable state space that the agent would not be able to efficiently explore.
\begin{figure}
    \includegraphics[width=8cm, height=6cm]{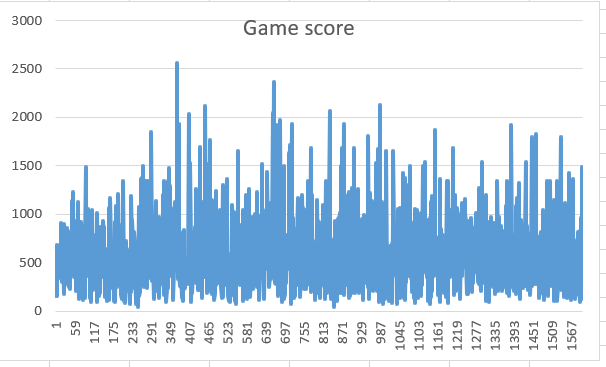}
    \caption{Two player implementation with trained assistant scores}
\end{figure}

 \begin{figure}
    \includegraphics[width=8cm, height=6cm]{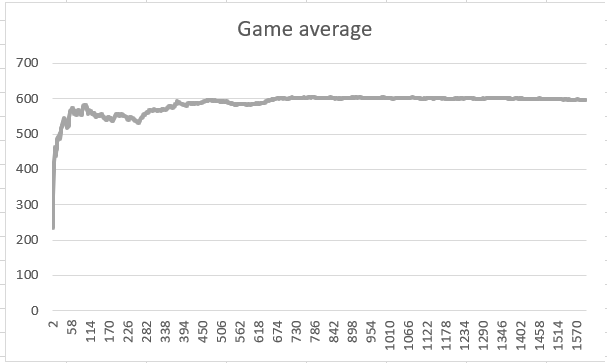}
    \caption{Two player implementation with trained assistant score average}
\end{figure}
The responses we obtained from those who played the game supports our interpretations of the observational data. Nearly all players agreed that the agent appeared to act with purpose rather than performing actions. The majority consensus is also that the agent appears more competitive than as an assistant which points to what we interpreted earlier about the agent performing the optimal actions. The reward function would need to be further modified to obtain the behavior we are aiming for. However the overall response is that the agent does help with winning the game and is an improvement of the gaming experience, except for one person whose dissatisfaction with the game experience has more to do with the assistant being invincible thus reducing the players agency to win the game as the agent could technically win by itself. This however has more to do with the implementation of the game mechanics rather than the agent itself.

\section{Future Works}
There are still areas for improvements in our experiment. Firstly, though the observations indicates that the version with the trained assistant shows better natural behavior than a random agent, a better reward mechanism can be applied to the assistant training step so that we can overcome the limitations mentioned before. Secondly, more training steps should be done to get a better behaved assistant agent. Since there are two players in the game, the number of the possible states in the game is squared, which requires much more training steps than we would for training a single-player agent to get good results. Last, but not least, searching through some of the state does not give us any useful information, so maybe it is possible to add a heuristic factor inside the training algorithm, which will reduce the training time in a certain degree.

\section{Conclusion}
In this project we aim to implement an intelligent AI assistant using an already successful reinforcement learning technique but altering the rewards. The results that we have collected for the single player agent training shows that the agent is able to play optimally to a decent level. It even employs tactics like shooting through the blockers such that the majority of its body is defended by the blocker while still being able to shoot enemies through the gap it creates. We have obtained results that show the agent can perform as a suitable assistant that improves the final score and serves as a crutch, but requires much more work from the developers part to act as a substitute for a human-player. But by being able to easily develop assistant agents that can help players progress through difficult stages developers can afford more leeway for level design and difficulty management. The success of this project also means that using reinforcement learning developers can develop agents self-taught in comparison to a rigid hard coded bot and speed up the development cycle. However deep Q-learning itself is only suited for simple 2D environments and would be very ineffective at dealing with 3D environments. But with the more progress in algorithms such like the A3C algorithms aimed towards 3D environments we would be able to replicate the results we obtained here in 3D games.

\section{References}
Papers: \newline \newline
[1] Playing Atari with Deep Reinforcement Learning. Volodymyr Mnih Koray Kavukcuoglu David Silver Alex Graves Ioannis Antonoglou Daan Wierstra Martin Riedmiller. DeepMind Technologies. arXiv:1312.5602v1
\newline  \newline
[2] Asynchronous methods for deep reinforcement learning. Volodymyr Mnih, Adrià Puigdomènech Badia, Mehdi Mirza, Alex Graves, Timothy P. Lillicrap, Tim Harley, David Silver, Koray Kavukcuoglu
arXiv:1602.01783
\newline \newline
[3] Silver, D., Huang, A., Maddison, C. et al. Mastering the game of Go with deep neural networks and tree search. Nature 529, 484–489 (2016).
\newline \newline
[4] Silver, D., Schrittwieser, J., Simonyan, K. et al. Mastering the game of Go without human knowledge. Nature 550, 354–359 (2017).
\newline \newline
[5] Egorov, Maxim. "Multi-agent deep reinforcement learning." CS231n: Convolutional Neural Networks for Visual Recognition (2016).
\newline \newline
[6] Deep Reinforcement Learning with Double Q-Learning Hado van Hasselt , Arthur Guez, and David Silver. Google DeepMind. AAAI Publications, Thirtieth AAAI Conference on Artificial Intelligence
\newline \newline
[7] CraftAssist: A Framework for Dialogue-enabled Interactive Agents
Jonathan Gray, Kavya Srinet, Yacine Jernite, Haonan Yu, Zhuoyuan Chen, Demi Guo, Siddharth Goyal, C. Lawrence Zitnick, Arthur Szlam
arXiv:1907.08584 [cs.AI]
Weblinks

\end{document}